\documentclass[10pt,twocolumn,letterpaper]{article}

\usepackage[pagenumbers]{cvpr} 

\usepackage{graphicx}
\usepackage{amsmath}
\usepackage{amssymb}
\usepackage{booktabs}

\usepackage[pagebackref,breaklinks,colorlinks]{hyperref}

\usepackage[capitalize]{cleveref}
\crefname{section}{Sec.}{Secs.}
\Crefname{section}{Section}{Sections}
\Crefname{table}{Table}{Tables}
\crefname{table}{Tab.}{Tabs.}

\begin{document}

\title{BeautyREC: Robust, Efficient, and Component-Specific Makeup Transfer}
\author{Qixin Yan$^1$\quad  Chunle Guo$^2$\quad  Jixin Zhao$^3$\quad  Yuekun Dai$^3$\quad Chen Change Loy$^3$\quad  Chongyi Li$^3$\footnotemark[1] \\
	\small{$^1$} \small WeChat, Tencent \quad
	\small{$^2$} \small TMCC, CS, Nankai University\quad	
	\small{$^3$} \small S-Lab, Nanyang Technological University \quad\\
	{\tt\small qixinyan@tencent.com \quad guochunle@nankai.edu.cn \quad zhao0388@e.ntu.edu.sg}\\
    {\tt\small  ydai005@e.ntu.edu.sg \quad ccloy@ntu.edu.sg \quad chongyi.li@ntu.edu.sg}\\
	{\tt\small \url{https://li-chongyi.github.io/BeautyREC_files}}
}
\maketitle
\thispagestyle{empty}
\renewcommand{\thefootnote}{\fnsymbol{footnote}}
\footnotetext[1]{Chongyi Li (chongyi.li@ntu.edu.sg) is the corresponding author.}

\begin{abstract}
In this work, we propose a \textbf{R}obust, \textbf{E}fficient, and \textbf{C}omponent-specific makeup transfer method (abbreviated as \textbf{BeautyREC}).
A unique departure from prior methods that leverage global attention, simply concatenate features, or implicitly manipulate features in latent space,  we propose a component-specific correspondence to directly transfer the makeup style of a reference image to the corresponding components (e.g., skin, lips, eyes) of a source image, making elaborate and accurate local makeup transfer.
As an auxiliary, the long-range visual dependencies of Transformer are introduced for effective global makeup transfer.
Instead of the commonly used cycle structure that is complex and unstable,  we employ a content consistency loss coupled with a content encoder to implement efficient single-path makeup transfer.
The key insights of this study are modeling component-specific correspondence for local makeup transfer,   capturing long-range dependencies for global makeup transfer, and enabling efficient makeup transfer via a single-path structure. 

We also contribute \textbf{BeautyFace}, a makeup transfer dataset to supplement existing datasets. 
This dataset contains 3,000 faces, covering more diverse makeup styles, face poses, and races.
Each face has annotated parsing map.
Extensive experiments demonstrate the effectiveness of our method against state-of-the-art methods.
Besides, our method is appealing as it is with only 1M parameters, outperforming the state-of-the-art methods (BeautyGAN: 8.43M,  PSGAN: 12.62M, SCGAN: 15.30M, CPM: 9.24M, SSAT: 10.48M).
\end{abstract}

\section{Introduction}
Makeup transfer is the problem of transferring the makeup style from a reference image to a source image without changing the identity and non-makeup regions of the source image.
Virtual makeup applications allow people to find well-suited makeup styles online or from reference images. 
More and more software companies and cosmetics companies  pay attention to the development of customized makeup transfer.

Transferring makeup style between a source image and a reference image is challenging. 
The large misalignment is easy to lead to the makeup leak.
The use of real paired training data is almost impossible. 
The identity and non-makeup regions of source image are fragile in the process of unsupervised learning, resulting in distorted textures and artifacts in the  result.
Efficient makeup transfer is also demanded as transfer algorithms are usually deployed on resource-limited and real-time devices such as mobile platforms.

To deal with these challenging issues, numerous methods \cite{Pairedcyclegan,BeautyGlow,SCGAN,LADN,PSGAN,BeautyGAN,SOGAN,Nguyen} have been proposed in recent years.
In addition to the traditional method \cite{GuoCVPR2009} that employs layer decomposition and layer-aware makeup transfer, contemporary methods mainly adopt unsupervised GANs to learn the transfer.
This is because acquiring sufficient real paired makeup/non-makeup training data is impractical. 
While existing methods are capable of transferring makeup, they have some inherent shortcomings. 
Some methods \cite{Pairedcyclegan,BeautyGlow,BeautyGAN} cannot cope with spatial misalignment between the reference image and the source image well due to the simple concatenation of source and reference features.
Preserving the identity and non-makeup regions of the source image after makeup transfer is crucial for a good user experience; however, some methods \cite{BeautyGlow,SCGAN,LADN,PSGAN} neglect this key fact in their designs.
As a result, they cannot achieve accurate makeup transfer and even damage the non-makeup background regions.
Additionally, existing methods adopt complex cycle network structures. Hence, they inevitably lead to a high memory footprint and long inference time, which constrain their practical applications. For instance, the trainable parameters of state-of-the-art methods are BeautyGAN \cite{BeautyGAN}: 8.43M,  PSGAN \cite{PSGAN}: 12.62M, SCGAN \cite{SCGAN}: 15.30M, CPM (color subnet only) \cite{Nguyen}: 9.24M and SSAT \cite{sun2022ssat}: 10.48M. 

\begin{figure*}[!t]
	\centering
	\centerline{\includegraphics[width=0.99\linewidth]{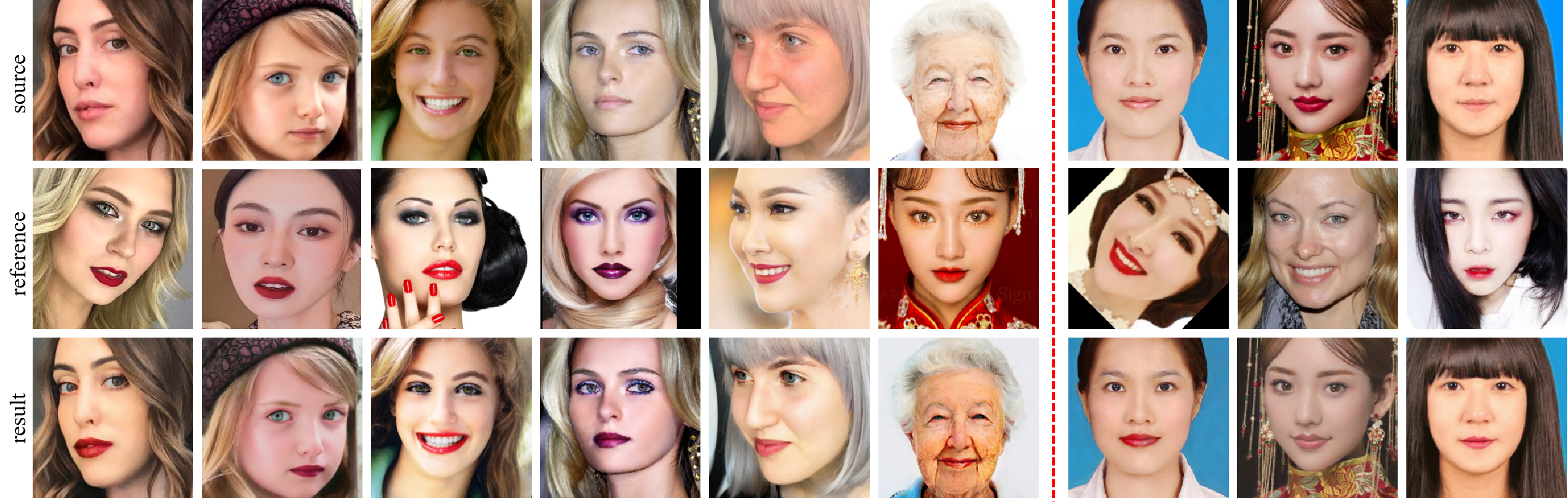}}
	\vspace{-0.5em}
    \caption{\textbf{A set of examples of \textit{BeautyREC}}. \textbf{Left}: Our method can effectively transfer diverse makeup styles, handle different ages, and preserve the identity and non-makeup regions of the source images. \textbf{Right}: Our method can implement flexible makeup transfer, such as handling the large spatial misalignment between the reference image and the source image, achieving the makeup removal by swapping the source image and the reference image,  and producing the component-specific makeup transfer (only lips).}
    \label{teaser}
    \vspace{-0.5em}
\end{figure*}

In this paper, we propose a \textbf{R}obust, \textbf{E}fficient, and \textbf{C}omponent-specific makeup transfer method, abbreviated as \textbf{BeautyREC}, to overcome the aforementioned issues.
Instead of leveraging global attention \cite{SOGAN}, simply concatenating source features and reference features \cite{Pairedcyclegan,BeautyGlow,BeautyGAN} or implicitly manipulating the component features in latent space \cite{SCGAN}, we devise a component-specific correspondence together with the corresponding component-specific discriminators to elaborately transfer the makeup styles of different components (e.g., skin, lips, eyes) in the reference image to the corresponding components of the source image. 
This not only avoids the artifacts induced by spatial misalignment but also preserves the non-makeup regions of the source image well. 

The most related work to our component-specific correspondence is the part-specific style encoder of SCGAN \cite{SCGAN}. 
Both works use the parsing maps to extract the component features. 
Different from SCGAN which implicitly maps the component features into an intermediate latent space and fuses them with the source features by a fusion block, our method explicitly transfers the component-specific makeup to the source image via our component-specific correspondence.
Besides, unlike SCGAN which discards the spatial information of makeup features by encoding them into a style code, we use spatial information of makeup features, which benefits the transfer of spatial makeup style such as the cheek color.
Note that the parsing maps are commonly used in makeup transfer methods and the current parsing map estimation and semantic segmentation methods \cite{BiSeNet} are stable in most cases.
To achieve more effective global makeup transfer, we use a Transformer-based structure, as an auxiliary of the component-specific correspondence, to capture the long-range visual dependencies between the source image and reference image.
Such a synergy of local and global makeup transfer cannot be achieved by previous methods. 

To preserve the image content of the source image, all existing makeup transfer methods adopt complex CycleGAN \cite{cyclegan} structures that introduce the cycle consistency loss to convert the image between the source domain and the reference domain.
However, we found that a content consistency loss that constrains the content similarity between the transferred image and the source image in the feature space, coupled with a content encoder, could implement an efficient single-path makeup transfer network and preserve the content of the source image well.
Consequently, the cycle structures for makeup transfer are no more required.
Our model has only about 1M parameters, which outperforms state-of-the-methods by a large margin.
\textit{We wish to emphasize that 
it is non-trivial to make a makeup transfer model such efficient.}

In Fig.~\ref{teaser}, we show a set of visual results of our method. 
As shown, the transfer of diverse makeup styles and large spatial misalignment between the reference images and the source image is made possible in our method. Moreover, our method effectively preserves the identity and non-makeup regions of the source images. 
Apart from the robustness, accuracy, and efficiency, our method also supports diverse applications such as makeup removal and component-specific makeup transfer.
Notably, these flexible applications are implemented with a single model. 

To facilitate the research on makeup transfer, we contribute a new makeup transfer dataset, \textbf{BeautyFace}, to supplement existing datasets.
In comparison to existing datasets, our dataset contains more diverse makeup styles, face poses, and
races. 
Accompanied with each face image, we also provide its parsing map.

Our main contributions are summarized as follows. 
\textbf{(1)} We propose a component-specific correspondence for accurate component-to-component makeup transfer. 
\textbf{(2)}  We propose a Transformer-based global makeup transfer, which models the long-range visual dependencies between the reference image and the source image.  
\textbf{(3)}  We employ a content consistency loss coupled with a content encoder, which endows our method with an extremely lightweight makeup transfer structure.
\textbf{(4)}  We contribute a new makeup transfer dataset, containing more diverse faces and the corresponding parsing annotations.

\begin{figure*}[!t]
	\centering
	\centerline{\includegraphics[width=0.9\linewidth]{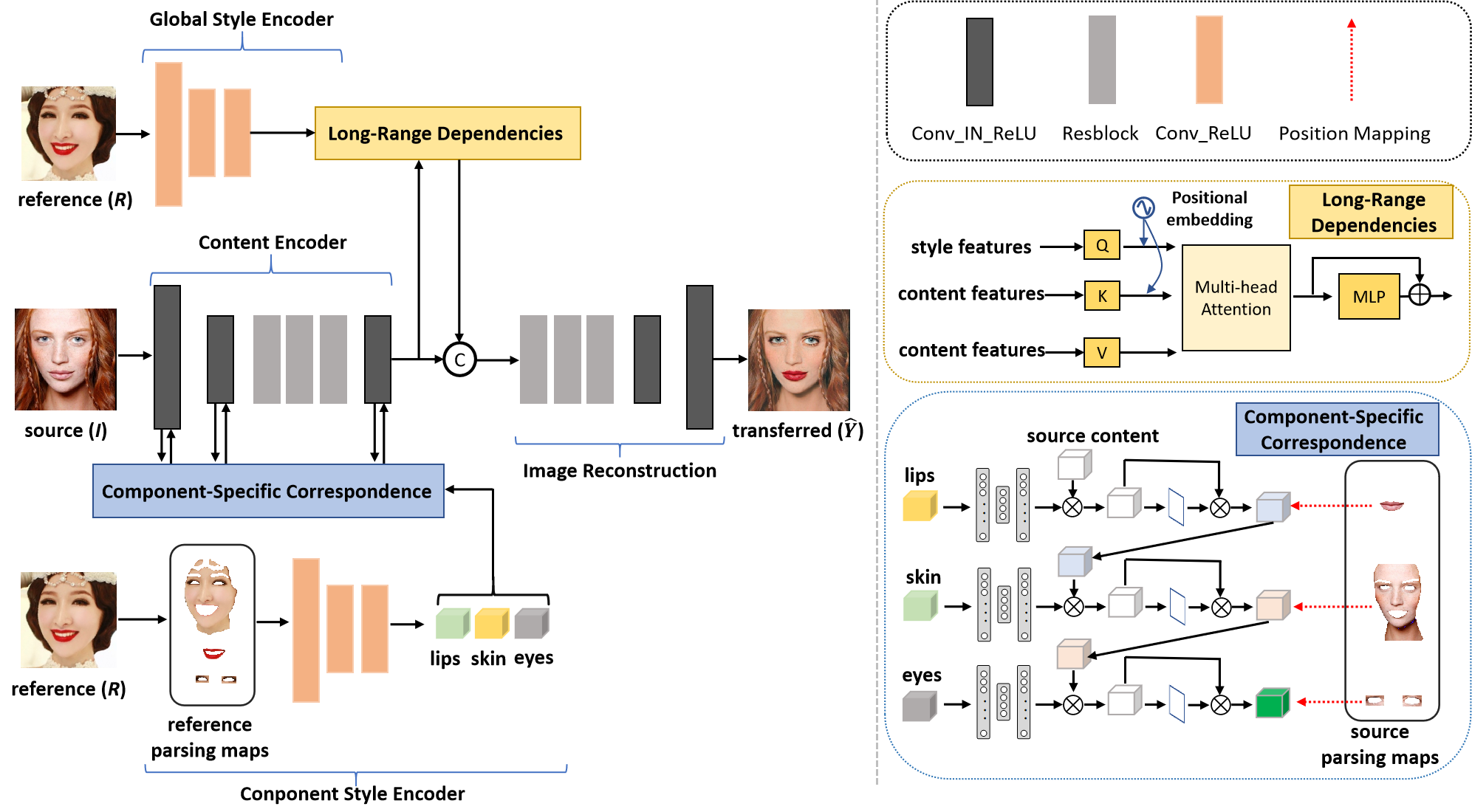}}
	\caption{\textbf{Overview of BeautyREC.} It consists of a content encoder, a component style encoder, a global style encoder, a component-specific correspondence, a long-range dependency, and an image reconstruction. Note that the skip-connections between the content encoder and image reconstruction are removed in figure for brevity.}
	\label{fig:framework}
\end{figure*}

\section{Related Work}
Traditional makeup transfer methods mainly employ  layer decomposition \cite{GuoCVPR2009} and face landmarks detection \cite{XuICIP2013} to transfer the makeup of an example to a source image.

Recently, deep learning has been widely used in makeup transfer, especially focusing on the usage of unsupervised GANs \cite{Pairedcyclegan,BeautyGlow,SCGAN,LADN,PSGAN,BeautyGAN,SOGAN,Nguyen,sun2022ssat,yang2022elegant}. 
All these methods adopt Cycle-GAN \cite{cyclegan}  structures to preserve the content of the source image. 
However, cycle structures need more training time and are unstable to preserve the identity and non-makeup regions of the source image. 
For example, BeautyGAN \cite{BeautyGAN} employed cycle consistency loss, perceptual loss, adversarial loss, and makeup loss to train the makeup transfer network. 
To compute the makeup loss, the parsing maps of the source image and reference image are used.
PSGAN \cite{PSGAN} was proposed to improve the robustness of makeup transfer for the large pose and expression differences between the reference image and the source image. 
In PSGAN, the makeup of the reference image was disentangled as two spatial-aware makeup matrices that were used to modify the features of the source image for achieving the corresponding makeup style. 
However, it is difficult to accurately transfer the makeup style to the source image with such global attention. 
The parsing maps of the source image and reference image are also needed to compute the makeup loss in the PSGAN.
To overcome the limitation of PSGAN  that uses global attention, SCGAN \cite{SCGAN} separately extracted the part-specific style features from the reference image and encoded them into a style-code in an intermediate latent space.
The parsing maps of the reference image are needed for extracting the part-specific style features. 
The part-specific style code is fused with the identity code via a makeup fusion block. 
Although SCGAN  can extract the part-specific features, the feature representations of each part are vague in latent space.   
Moreover, the feature fusion process of SCGAN only considers the statistic information (i.e., using AdaIN \cite{AdaIn}) but neglects the spatial style.
SOGAN \cite{SOGAN} proposed a  shadow and occlusion robust method for makeup transfer, in which the reference image and the source image are combined in the UV space.
EleGANt \cite{yang2022elegant} utilized high-resolution feature maps to preserve high-frequency makeup features beyond color distributions. 
SSAT \cite{sun2022ssat} proposed a semantic-aware Transformer network and a weakly supervised semantic loss to achieve semantic correspondence.

As another line, Nguyen et al. \cite{Nguyen} focused on both makeup color transfer and pattern transfer. 
For makeup color transfer, this method follows the traditional cycle structure.
To implement pattern transfer, a parallel branch with the makeup color transfer  branch is used to estimate the 
pattern mask that copes the pattern on the reference image and pastes it on the source image.
Similar to the majority, the focus of our study is to transfer the makeup styles, excluding the pattern transfer.
Thus, we only compare our method with the makeup color transfer branch of  Nguyen et al.'s method.


\section{BeautyREC}
\subsection{Network Structure}

\noindent
\textbf{Overview.} 
BeautyREC is a single-path structure, as illustrated in Fig. \ref{fig:framework}. 
First, the content features of the source image are extracted by the content encoder. 
With the use of content consistency loss in feature space, the content features are insensitive to the makeup of the source.
More discussions are provided in the Ablation Study.
Second, a global style encoder is used to obtain the global makeup style of the reference image while a component style encoder aims to extract the style features of different components of the reference image.
Third, with the features of lips style, skin style, and eyes style of the reference image, we transfer them to the corresponding component of the source image using a component-specific correspondence. 
Fourth, with the global features of the reference image, the long-range visual dependencies between the reference image and the source image are modeled by the multi-head self-attention.
At last, image reconstruction is employed to integrate features and produce a makeup transferred image.
We provide detailed network structure and parameters in the supplementary material.

\noindent
\textbf{Content Encoder.} 
In the practical applications of makeup transfer algorithms, the source image is usually covered by makeup, which increases the difficulty of transferring makeup from a reference image to the source image.
This also may lead to the makeup overlay in the final result. 
However, this issue is commonly neglected in previous methods. 
Thus, they prefer the non-makeup source image in the inference process. 
To cope with this issue, we use a content encoder together with a content consistency loss in feature space to make the content encoder features of the source image insensitive to makeup style. 
The content consistency loss will be detailed in the Objective Function.

To achieve an efficient network, the content encoder contains only three Conv-IN-ReLU layers and three Resblocks, as shown in Fig. \ref{fig:framework}. Each Resblock includes two convolution layers with a residual connection. We downsample the features of the first Conv-IN-ReLU layer.

\noindent
\textbf{Style Encoder.} 
We adopt the same network structure to extract the component style features (i.e., component style encoder) and the global style features (i.e., global style encoder), respectively.
The difference between the two style encoders is that the component style encoder separately extracts the features of different components using the corresponding parsing map, producing the component-specific features.
The component style encoder also endows the flexible controllability of component-specific makeup transfer.  

Specifically, we first binarize the parsing maps $R_{par}$ of the reference image $R$, i.e., setting the corresponding component region (skin, lips, or eyes) to 1 and other regions to 0. 
Then, the corresponding component  of the reference image is obtained by
\begin{equation}
	\label{equ_2}
	R^{com}=R \odot \tilde{R_{par}^{com}}, com \in\{skin,lips, eyes\},
\end{equation}
where $R^{com}$ represents the component $com$ of the reference image, $\tilde{R_{par}^{com}}$  represents a  binarized parsing map,  and $\odot$ is the Hadamard product. 
$R^{com}$ is separately fed to the three Conv-ReLU layers to achieve the corresponding component's makeup features. 
After the first Conv-ReLU layer, a downsampling operation is followed.

\noindent
\textbf{Component-Specific Correspondence.} 
We propose a component-specific correspondence to perform accurate makeup transfer for different components, taking the statistic information and spatial information of makeup style into account in the makeup transfer process.

As shown in the bottom right corner of Fig. \ref{fig:framework}, with three sets of component-specific makeup features, the content features of the source image go through a \textit{component-to-component} (from the reference's component to the source's component) transfer and a \textit{component-by-component} (from lips, skin, to eyes in the source image) transfer. 
Following the arrows, we transferred the lips, skin, and eyes styles of the reference image one by one to the same components of the source image according to the corresponding semantic parsing map of  the source image.
Taking the skin style transfer as an example, the source content used here is the lips style transferred features.
We first use channel attention to scale the features of content features from a statistical perspective. 
Then, we further process the scaled features by spatial attention. 
To accurately transfer the specific component's makeup style to the corresponding region of the source image, we adopt the position mapping to only change the features in the corresponding region.  
The process can be formulated as:
\begin{equation}
	\label{equ_3}
	F_{trans}^{lips}=\text{PM}(\text{SA}(\text{CA}(F_{style}^{lips})\otimes F_{cont})),
\end{equation}
\begin{equation}
	\label{equ_22}
	F_{trans}^{skin}=\text{PM}(\text{SA}(\text{CA}(F_{style}^{skin})\otimes F_{trans}^{lips})),
\end{equation}
\begin{equation}
	\label{equ_4}
	F_{trans}^{eyes}=\text{PM}(\text{SA}(\text{CA}(F_{style}^{eyes})\otimes F_{trans}^{skin})),
\end{equation}
where  $F_{tra}^{lips}$, $F_{trans}^{skin}$, and 	$F_{tra}^{eyes}$  are the only lips transferred features, lips and skin transferred features, and lips, skin, and eyes transferred features.
$F_{style}^{lips}$, $F_{style}^{skin}$, and $F_{style}^{eyes}$ are the style features that refer to the lips, skin, and eyes of the reference image.
$F_{cont}$ denotes the content features of the source image. 
$\otimes$ represents the pixel-wise multiplication.
$\text{PM}$, $\text{SA}$, and $\text{CA}$ represent the position mapping, spatial attention, and channel attention, respectively.

In the position mapping $\text{PM}$, we first compute the binarized parsing map $\tilde{I_{par}^{com}}$ of the source image $I$. 
Then, the output $F^{com}_{PM}$ of the position mapping can be expressed as:
\begin{equation}
	\label{equ_5}
	F^{com}_{PM}=F_{SA} \odot \tilde{I_{par}^{com}}\oplus F_{in} \odot (1-\tilde{I_{par}^{com}}),
\end{equation}
where $F_{SA}$ denotes the output features of the corresponding spatial attention, $F_{in}$  represents the input features of the corresponding channel attention, and $\oplus$ represents the pixel-wise addition.
In this way, we only transfer the components' makeup styles of the reference image to the source image's corresponding components, thereby avoiding damaging the non-makeup regions of the source image.

\begin{figure*}[!t]
	\centering
	\centerline{\includegraphics[width=0.85\linewidth]{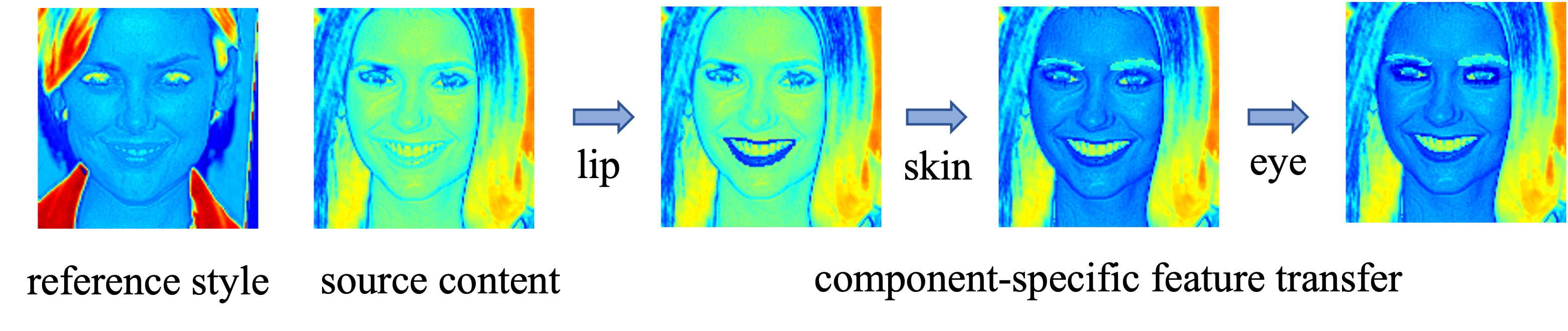}}
    \vspace{-1.5em}
	\caption{\textbf{Visualization of feature transfer process in our component-specific correspondence.} We normalize the values of feature maps to the range of [0,1] and average them for visualization in heatmaps.}
	\label{CST}
\end{figure*}

In Fig. \ref{CST}, we show the feature changes in the process of component-specific feature transfer. 
As shown, the feature of source content changes in the specific regions from lips, skin, to eyes, using the component-specific correspondence between the reference style features and the source content features, thus implementing the component-specific feature transfer.
Additionally, it is insensitive to the order of feature transfer, which is discussed in the supplementary material.

\noindent
\textbf{Long-Range Dependencies.}
The component-specific correspondence may be insufficient for processing global makeup style transfer because of the inherent limitations of convolution layers such as the local modeling properties.
Thus, we employ a Transformer \cite{16words,Swin} to exploit long-range visual dependencies between the source image and reference image.

As illustrated in Fig. \ref{fig:framework}, the basic components of our Transformer are Query (Q): style features, Key (K): content features, and Value (V): content features. 
It is intuitive that the style features are used as Query to model the dependencies with content features (Key).
The Query is used to model the long-range visual dependencies with the Key via multi-head attention \cite{MHAttention}. 
The use of multi-head attention allows our network to jointly attend the information from different representation spaces of different positions.
With the obtained long-range dependencies between Query and Key, the purpose is to transfer the makeup to the Value globally.
With the attention maps, the Value is weighted.
The weighted Value goes through an MLP and then produces the output features. 
The process can be expressed as:
\begin{equation}
	\label{equ_7}
	F_{gstyle}=\text{MLP}(\text{MHA}(Query, Key, Value;Pos)),
\end{equation}
where $F_{gstyle}$ represents the output features of global transfer, $\text{MLP}$ is a two-layer MLP with a residual connection, $\text{MHA}$ is a multi-head attention module with eight heads, and $Pos$ is the  sine and cosine-based position embedding.

\noindent
\textbf{Image Reconstruction.}
Based on the makeup style transferred features from the component-specific transfer and the global transfer, we employ an image reconstruction to refine the features and recover the image resolution. 
The image reconstruction has a symmetrical structure with the content encoder, in which the component-specific transferred features and the global features are concatenated with the corresponding decoder features. 

\subsection{Objective Function}
\label{sec:loss}

\noindent
\textbf{Content Consistency Loss.}
To reduce the effect of the makeup of source image on makeup transfer performance and preserve the content of source image in the transferred result, we constrain the content consistency in feature space:
\begin{equation}
	\label{equ_6}
	\mathcal{L}_{\text{cont}}=\left \|\theta_{\text{cont}}(I)-\theta_{\text{cont}}(\hat{Y})\right\|,
\end{equation}
where $I$ represents the source image, $\hat{Y}$ represents the transferred result, $\mathcal{L}_{\text{cont}}$ denotes the content consistency loss, $\theta_{\text{cont}}$ is the first Conv-IN-ReLU layer of our content encoder.

\noindent
\textbf{Makeup Loss.}
Following previous methods \cite{SCGAN,BeautyGAN}, our method also uses the makeup loss  that consists of local histogram matching on different components of the source images and the reference image:
\begin{equation}
	\label{equ_8}
	\begin{aligned}
		\mathcal{L}_{\text{mu}}=&\left \| G(I, R)-\text{HM} (I,R)\right\|_{2}+\left \| G(R, I)-\text{HM} (R,I)\right\|_{2},
	\end{aligned}
\end{equation}
where  $\mathcal{L}_{\text{mu}}$ denotes the makeup loss, $R$ represents the reference image, $G$ denotes our makeup transfer network, and $\text{HM}(\cdot)$ is the histogram matching.

\noindent
\textbf{Perception Loss.}
To preserve the perception similarity between source image and transferred result, we denote the perception loss  $\mathcal{L}_{\text{per}}$ as:
\begin{equation}
	\label{equ_9}
	\mathcal{L}_{\text{per}}=\left \|\theta_{\text{vgg}}(I)-\theta_{\text{vgg}}(\hat{Y})\right\|_{2},
\end{equation}
where $\theta_{vgg}$ represents the pre-trained VGG-19 network \cite{VGG}. We use the conv4 layer before the activation function.

\noindent
\textbf{Adversarial Loss.}
In addition to the global adversarial loss $\mathcal{L}_{\text{ad}}^{\text{global}}$, we also employ local adversarial losses to further enhance the significance of the local makeup style. 
Thus, our method is equipped with five discriminators, including a global discriminator, a skin discriminator, a lips discriminator, a left eye discriminator, and a right eye discriminator. The final adversarial loss can be expressed as:
\begin{equation}
	\label{equ_10}
	\mathcal{L}_{\text{ad}}=\mathcal{L}_{\text{ad}}^{\text{global}}+\mathcal{L}_{\text{ad}}^{\text{skin}}+\mathcal{L}_{\text{ad}}^{\text{lips}}+\mathcal{L}_{\text{ad}}^{\text{leye}}+\mathcal{L}_{\text{ad}}^{\text{reye}},
\end{equation} 
where $\mathcal{L}_{\text{ad}}^{\text{skin}}$, $\mathcal{L}_{\text{ad}}^{\text{lips}}$, $\mathcal{L}_{\text{ad}}^{\text{leye}}$, and $\mathcal{L}_{\text{ad}}^{\text{reye}}$ are the local component-specific adversarial losses. 

The total loss is a combination of the above-mentioned losses, which can be expressed as:
\begin{equation}
	\label{equ_11}
	\begin{aligned}
		\mathcal{L}_{\text{total}}=\mathcal{L}_{\text{cont}}+\mathcal{L}_{\text{mu}}+\lambda_{\text{per}}\mathcal{L}_{\text{per}}+\lambda_{\text{ad}}\mathcal{L}_{\text{ad}},\\
	\end{aligned}
\end{equation}
where $\lambda_{\text{per}}$=0.005  and $\lambda_{\text{ad}}$= 0.5 are the corresponding weights for balancing the magnitudes of losses.

\subsection{BeautyFace}
\label{sec:dataset}
While there are some makeup datasets\cite{BeautyGAN,PSGAN}, their diversity is insufficient and resolution is low (commonly 256$\times$256).
Especially, some of them were collected several years ago, thus excluding the new fashion styles.
To supplement existing makeup datasets, we collect a new dataset from the Internet, named BeautyFace. 
It contains 3,000 high-quality face images with a higher resolution of 512$\times$512, covering more recent makeup styles and more diverse face poses, backgrounds, expressions, races, illumination, etc. 
Besides, we annotate each face with parsing, which benefits more diverse applications. 
We show some examples of BeautyFace in Fig. \ref{fig:dataset}. More results can be found in the supplementary material.

\begin{figure}[!h]
	\centering
	\centerline{\includegraphics[width=1\linewidth]{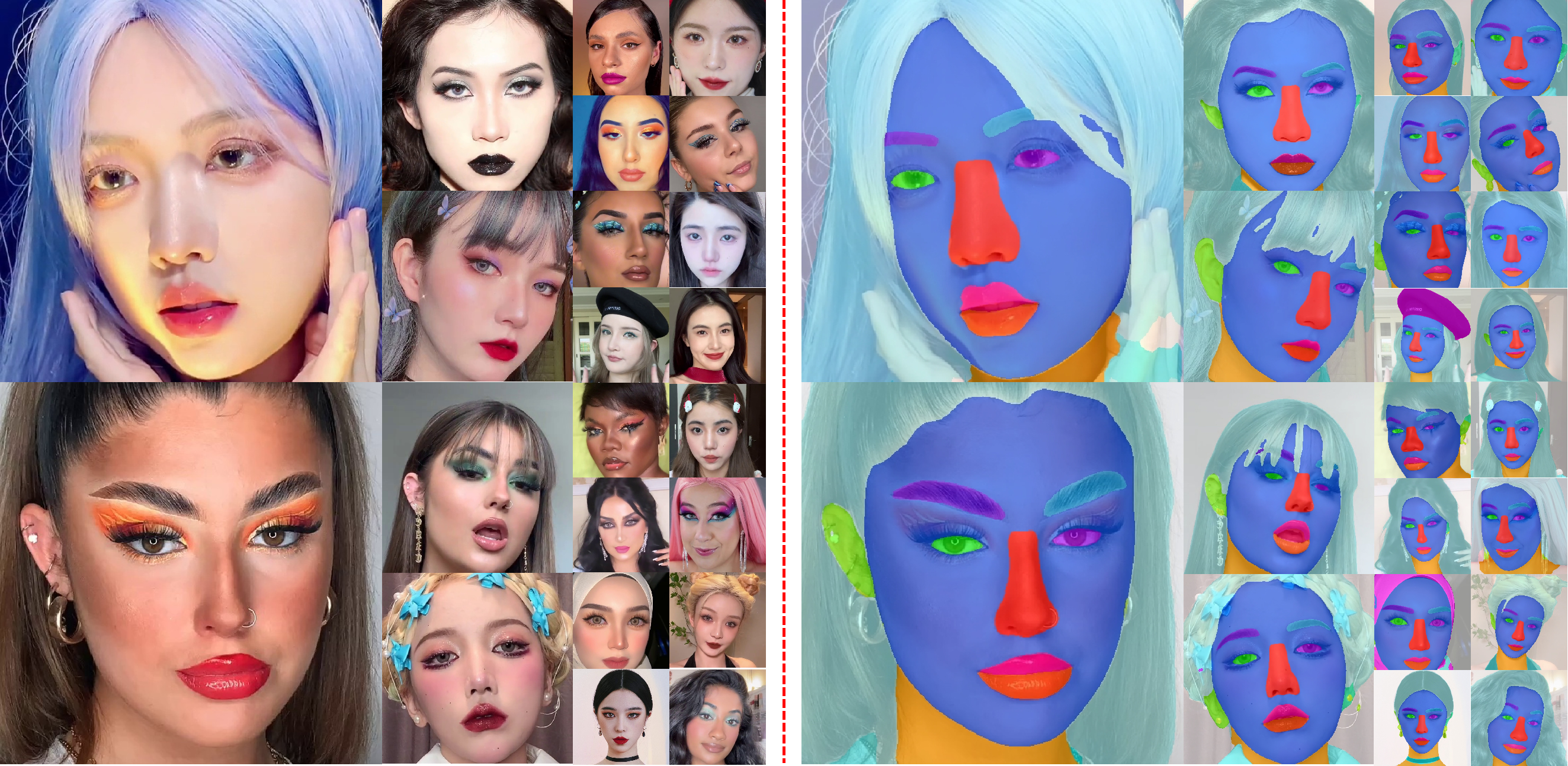}}
	\caption{\textbf{Examples of BeautyFace dataset.} We show several face images (left) and the parsing maps (right).}
	\label{fig:dataset}
	\vspace{-1em}
\end{figure}

\begin{figure*}[!t]
    \begin{center}
        \centering
        \includegraphics[width=1\textwidth]{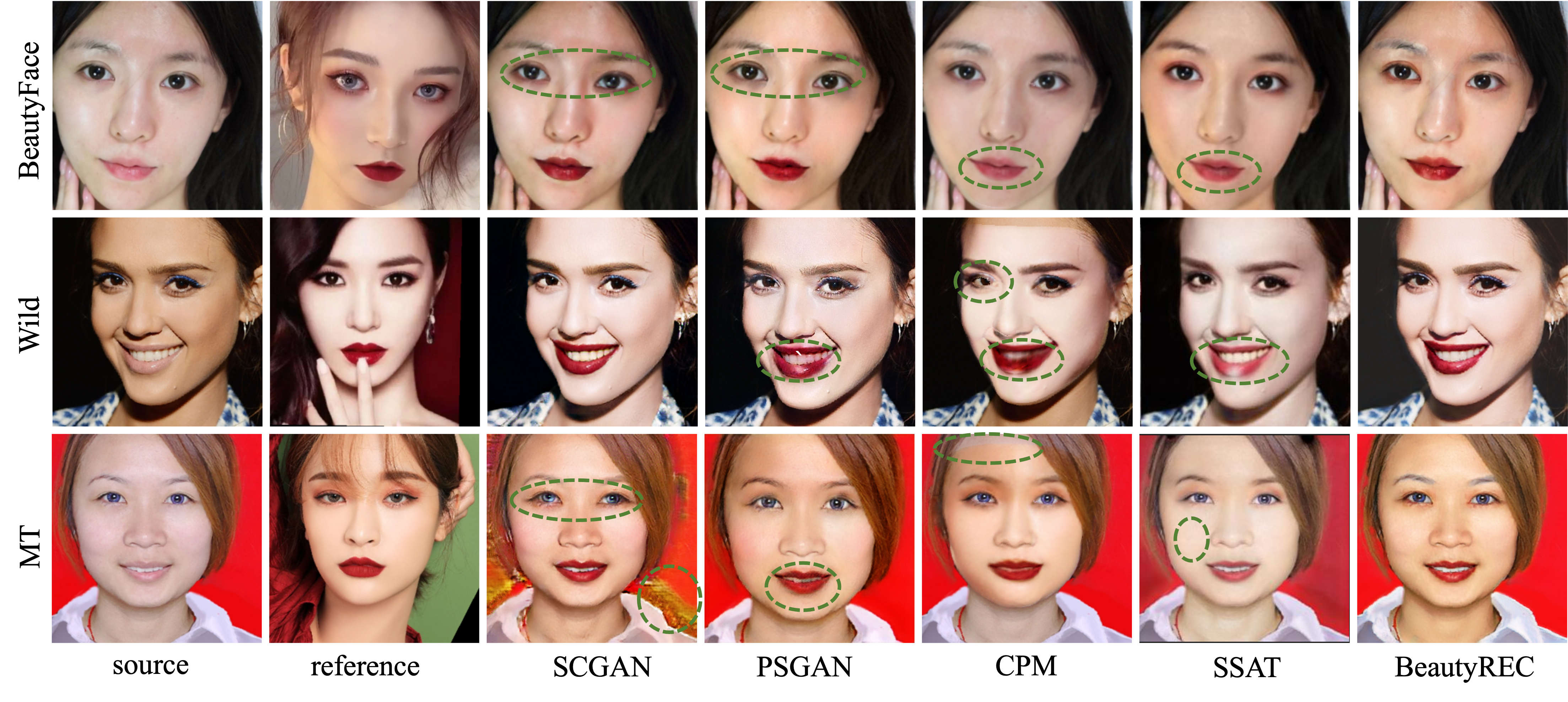}
        \vspace{-2.5em}
	\caption{\textbf{Visual comparison.} Compared with the state-of-the-art methods, our method successfully transfers the makeup from the reference images to the source images. It does not introduce makeup leak and artifacts and reserves the identity and non-makeup regions of the source images well. Zoom in for best view.}
        \label{fig:comparison}
    \end{center}%
    \vspace{-1em}
\end{figure*}


\section{Experiments}
\subsection{Experimental Settings}

\noindent
\textbf{Implementations.}	
BeautyREC is implemented with PyTorch	on an NVIDIA 2080Ti GPU. 
We train the model with an ADAM optimizer with the fixed learning rate 1 $\times$10$^{-4}$. 
The mini-batch size is set to 1.
For the discriminators used in our method, we adopt the U-Net discriminator \cite{wang2021realesrgan} for producing accurate gradient feedback for local regions. 
We follow previous methods \cite{SCGAN,BeautyGAN} to train our method on MT dataset \cite{BeautyGAN} that contains 3,834 images. 
We randomly select 100 non-makeup images and 250 makeup images for test. 
We also use the Wild \cite{PSGAN} and BeautyFace for test. 

\noindent
\textbf{Comparison Methods.}
We compare our method with several state-of-the-art makeup transfer methods: SCGAN \cite{SCGAN}, PSGAN \cite{PSGAN}, CPM \cite{Nguyen} and SSAT\cite{sun2022ssat}. We use the released code of these methods. 
We include only the released makeup color transfer model of CPM  for fair comparisons.
The pattern transfer of CPM is beyond the scope of this paper and existing makeup transfer methods.
Note that the code and pre-trained models of many makeup transfer methods are not publicly available.
Moreover, their training data are not clear for re-implementation. 
For fair comparisons, we only use the official models of different methods for experiments.

\begin{table}[!t]

     \vspace{-1em}
     \centering
     	\resizebox{8cm}{!}{
     \begin{tabular}{c|c|c|c|c|c}
     \hline 
        Datasets & SCGAN & PSGAN & CPM & SSAT & BeautyREC \\ \hline\hline
\multicolumn{6}{c}{ArcFace $\uparrow$}  \\ \hline
          Wild & 0.864&0.798&0.735& 0.831 & 0.883 \\

          MT & 0.857&0.829&0.767&0.892 &   0.878  \\ 
          BeautyFace & 0.879 & 0.849 & 0.919 & 0.835 & 0.893 \\ \hline
\multicolumn{6}{c}{Fid $\downarrow$} \\\hline
          Wild & 37.87 &33.51&41.30& 34.96 &  24.00 \\ 

          MT & 70.59&47.91&52.97& 37.33 &  38.14  \\ \hline  
          BeautyFace &52.29 &40.31 &109.06 & 38.33 & 32.79 \\\hline
     \end{tabular}
     }
          \caption{\textbf{Identity preservation comparisons on the MT, Wild, and BeautyFace testing sets.} }
     \label{table:fid}
     \vspace{-1em}
 \end{table}

\subsection{Experimental Comparisons}

\noindent
\textbf{Visual Comparisons.}
We first show several representative visual comparisons in Fig. \ref{fig:comparison}. 
As presented, the compared methods either produce unnatural makeup transfer or inaccurate transfer. For example, SCGAN, PSGAN, and CPM transfer the lipstick of the reference image to the teeth of the source image. 
SCGAN produces the artifacts in the regions of eyes.
CPM  introduces obvious artifacts in the transferred results. 
SSAT cannot effectively transfer the makeup from the reference images to the source images.
In addition, SCGAN  and SSAT cannot preserve the non-makeup regions well such as the washed-out background, and even damage the identity of the source images such as the eyes in the result of SCGAN. 
In comparison, our method not only effectively transfers the makeup style but also preserves the identity and non-makeup regions of the source images well.
More results could be found in the supplementary material.

\noindent
\textbf{Identity Preservation Comparisons.}
To compare the performance of different methods for identity preservation of source images, we calculate the average cosine similarity of ArcFace \cite{ArcFace} features between the faces before and after makeup transfer. 
We also use Fr${\acute{e}}$chet Inception Distance (FID) \cite{FID} that compares the distribution of transferred images with the distribution of source images. 
%
We randomly selected 100 pairs of images from the testing set of MT \cite{BeautyGAN}, 100 pairs of images from Wild dataset \cite{PSGAN}, 300 pairs of images from BeautyFace dataset for test. The results of identity preservation comparison are shown in Table \ref{table:fid}.
In Table \ref{table:fid}, our method achieves related good scores when compared with other methods.
Note that the ArcFace and FID cannot accurately reflect the makeup transfer performance as they may have good scores when the transferred result keeps the same as the source image, i.e., the transfer algorithm does not work. 
Following previous works, we only list the scores of these two metrics as reference.

\noindent
\textbf{User Study.}
Although some methods (CPM and SSAT) can obtain relatively good performance in some identity preservation comparisons, they have poor transfer quality. Hence, we perform a user study to quantify the visual quality of the transferred results. 
We randomly select 10 source images and 10 reference images from Wild and  10 reference images from BeautyFace, respectively,  and transfer the source images to the reference images using different methods.
For each set of transferred results, we invite 20 participants to independently rank them.
During ranking, these participants are trained by observing the results from 1) the makeup style similarity between the transferred result and the reference image; 2) the similarity of the identity and non-makeup regions between the transferred result and the source image; and 3) the realism of the transferred results such as artifacts and inappropriate color.
We present the best-selected ratio  in Table \ref{table:US}. 
Our method achieves the highest best-selected ratio, which suggests the better performance of our method for accurate makeup transfer than the compared methods.

\begin{table}[!t]
	\centering
	
	 \vspace{-1em}

	\begin{center}
 \resizebox{8cm}{!}{
		\begin{tabular}{c|c|c|c|c|c}
			\hline
	Methods &	PSGAN & CPM &SCGAN & SSAT &BeautyREC\\
			\hline\hline
		Ratio &	 4.5& 9.0 & 17.0 & 22.0  & \textbf{47.5}     \\
			\hline
		\end{tabular}
  }
	\end{center}

        \caption{\textbf{User study in terms of the best selected ratio (\%).}}
	\label{table:US}
	\vspace{-1em}
\end{table}

\noindent
\textbf{Model Size and Runtime Comparisons.}
We compare the model sizes and runtime in Table \ref{table:size}. 
Our method has the lowest trainable parameters and FLOPs and the fastest inference speed.
The results suggest the efficiency of our method.

	\begin{table}[!h]
		\centering

  
    \begin{center}
			\begin{tabular}{c|c|c|c}
				\hline
				Methods  & Parameters$\downarrow$ & FLOPs$\downarrow$ & runtime$\downarrow$ \\
				\hline\hline
				SCGAN          & 15.30  & 1154.46   & 0.1272      \\
				PSGAN         &  12.62 & 38.82  & 0.1005      \\
				CPM           &  9.24   &   66.89 &   0.1424     \\
				SSAT          & 10.48 & 737.24 & 0.0681  \\
				BeautyREC          &   \textbf{0.99}  & \textbf{12.59}  & \textbf{0.0236}      \\
				\hline
			\end{tabular}
		\end{center}
	
  		\caption{\textbf{Model size and runtime comparisons.} The trainable parameters (in M), FLOPs (in G), and runtime (in second) for processing a pair of source and reference images with a size of 256$\times$256 are computed.}
		\vspace{-1em}
		\label{table:size}
	\end{table}

\subsection{Ablation Study}
We conduct ablation studies to demonstrate the effectiveness of our novel designs, including the component-specific correspondence (CSC), long-range dependencies (LRD), and content consistency loss coupled with a content encoder ($\mathcal{L}_{\text{cont}}$). 
We retrain the ablated models while keeping the same settings as our method, except for the ablated parts.

We first conduct quantitative experiments in Table \ref{ablation}. As presented, the full model achieves the best identity preservation than the ablated models on Wild and MT datasets in terms of the ArcFace metric.

 \begin{table}[!h]

       \resizebox{8cm}{!}{
     \centering
     \begin{tabular}{c|c|c|c|c}
     \hline 
        Datasets & w/o $\mathcal{L}_{\text{cont}} $  & w/o CSC  & w/o LRD & full model \\ \hline\hline 
          Wild & 0.855 & 0.866 & 0.859 &\textbf{0.883} \\
          MT  & 0.842  & 0.838 & 0.861 &\textbf{0.878}  \\ \hline
     \end{tabular}
    \vspace{-1em}
     
     }
     
       \caption{\textbf{ArcFace ($\uparrow$) scores of the ablated models.} }
       \label{ablation}
 \end{table}

\begin{figure}[!h]
	\centering
	\centerline{\includegraphics[width=0.95\linewidth]{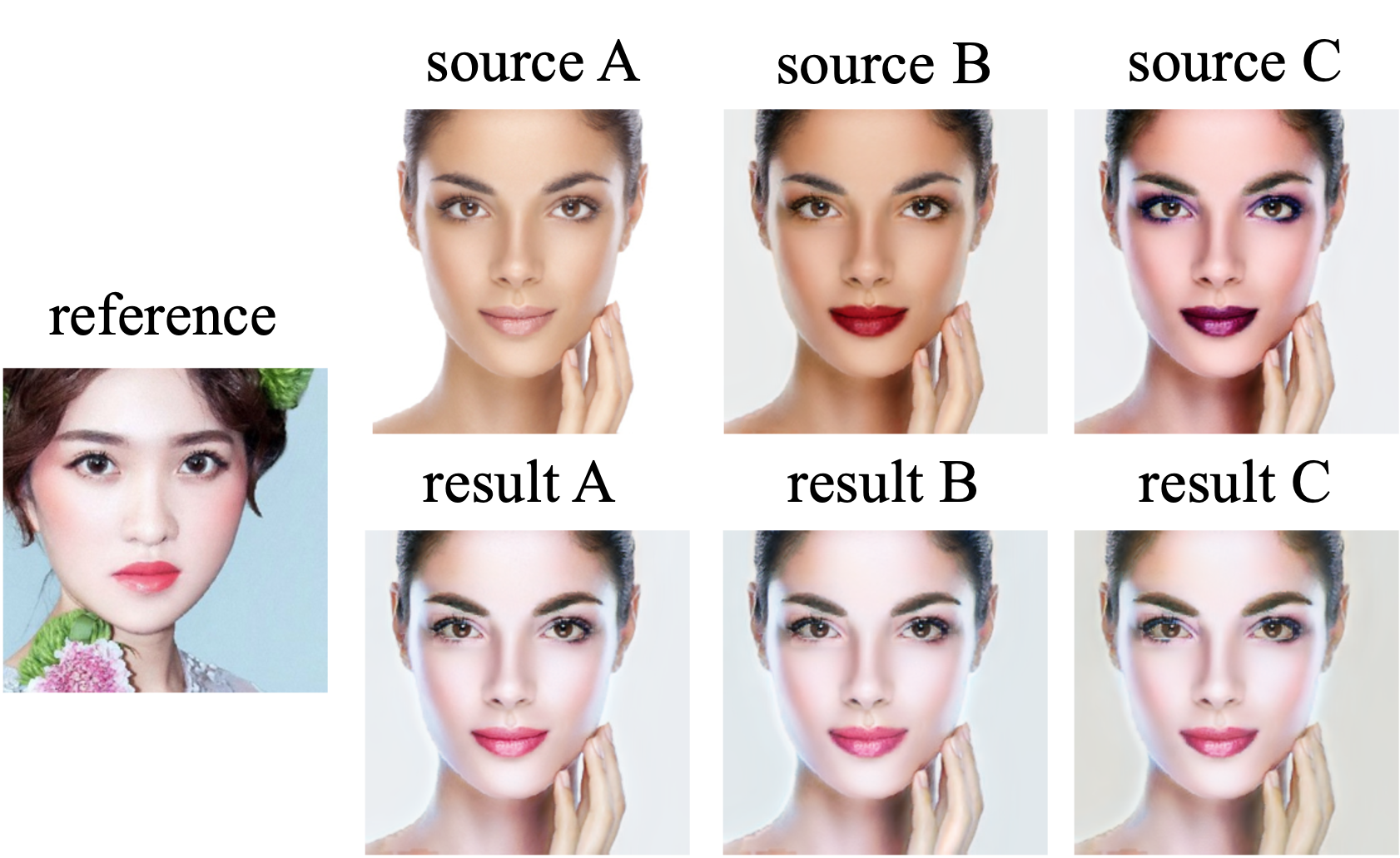}}
	\caption{\textbf{Robustness of our ReautyREC.}}
	\label{robust2}
	\vspace{-0.5em}
\end{figure}

\noindent
\textbf{Robustness of Our BeautyREC.} 
We provide the visual results in Fig. \ref{robust2}, which show the robustness of our method to the source image with or without makeup. As shown, our method can achieve the same transferred results regardless of whether the source image has makeup or not (The source B image and the source C image are covered by makeup while the source A image is not.). The results suggest that our method can eliminate the effect of the makeup on the source image,  benefiting from the content encoder together with the content consistency loss in feature space. 

\begin{figure}[!h]
	\centering
	\begin{minipage}[b]{0.23\linewidth}
		\centering
		\centerline{\includegraphics[width=1.8cm,height=1.8cm]{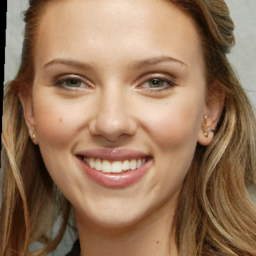}}
		\centerline{(a) source}\medskip
	\end{minipage}
	\begin{minipage}[b]{0.23\linewidth}
		\centering
		\centerline{\includegraphics[width=1.8cm,height=1.8cm]{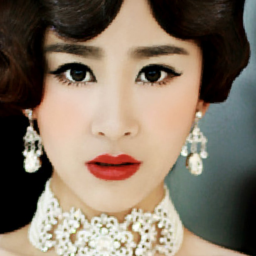}}
		\centerline{(b) reference}\medskip
	\end{minipage}
	\begin{minipage}[b]{0.23\linewidth}
		\centering
		\centerline{\includegraphics[width=1.8cm,height=1.8cm]{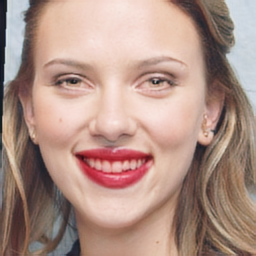}}
		\centerline{(c) w/o CST}\medskip
	\end{minipage}
	\begin{minipage}[b]{0.23\linewidth}
		\centering
		\centerline{\includegraphics[width=1.8cm,height=1.8cm]{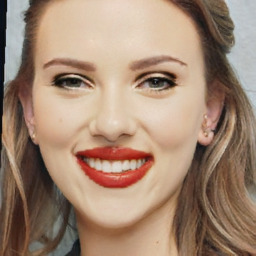}}
		\centerline{ (d)  w/ CST}\medskip
	\end{minipage}
	\vspace{-1em}

	\caption{\textbf{Effect of component-specific transfer.}}
	\label{fig:abs_cst}
 \vspace{-1em}
\end{figure}

\noindent
\textbf{Effectiveness of CSC.}
We replace the component-specific correspondence (CSC) with simple global attention (denoted as w/o CSC) that uses channel attention and spatial attention to globally modulate the source image features by the reference image features. Such global attention is commonly used in previous makeup transfer methods. 
As shown in Fig. \ref{fig:abs_cst}, the model-w/o CSC causes ambiguous makeup style transfer such as the teeth regions, and cannot transfer the local regions' makeup style well such as the eyes regions. 

\begin{figure}[!h]
	\centering
		\begin{minipage}[b]{0.23\linewidth}
		\centering
		\centerline{\includegraphics[width=1.8cm,height=1.8cm]{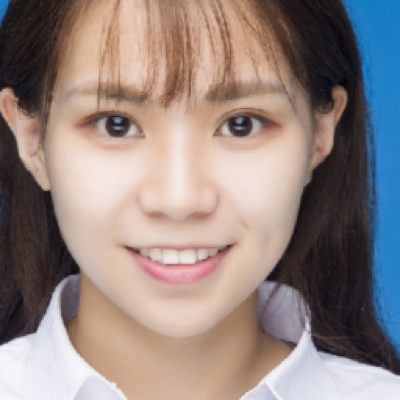}}
		\centerline{ (a) source}\medskip
	\end{minipage}
	\begin{minipage}[b]{0.23\linewidth}
		\centering
		\centerline{\includegraphics[width=1.8cm,height=1.8cm]{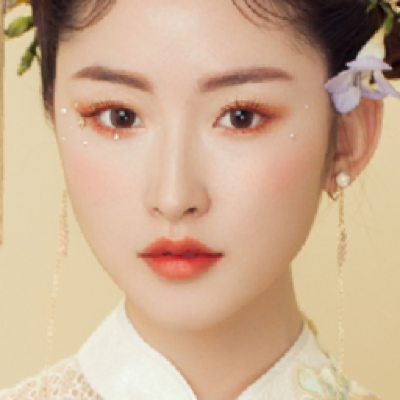}}
		\centerline{ (b) reference}\medskip
	\end{minipage}
	\begin{minipage}[b]{0.23\linewidth}
		\centering
		\centerline{\includegraphics[width=1.8cm,height=1.8cm]{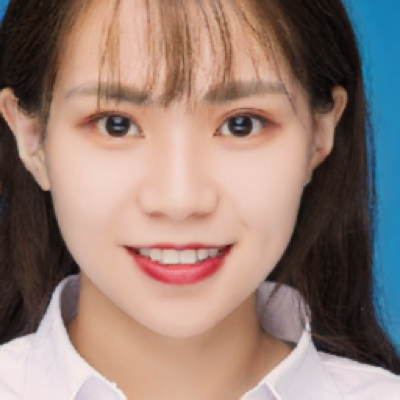}}
		\centerline{ (c) w/o LRD}\medskip
	\end{minipage}
	\begin{minipage}[b]{0.23\linewidth}
		\centering
		\centerline{\includegraphics[width=1.8cm,height=1.8cm]{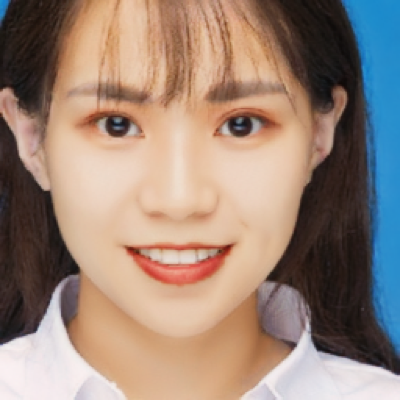}}
		\centerline{ (d) w/ LRD}\medskip
	\end{minipage}
    \vspace{-1em}
	\caption{\textbf{Effect of the long-range dependencies.}}
	\label{fig:abs_global}
	
\end{figure}

\noindent
\textbf{Effectiveness of LRD.}
We remove the long-range dependencies (LRD) and retrain the network. The comparison results are shown in Fig. \ref{fig:abs_global}. As shown, the use of long-range dependencies achieves better global makeup transfer.

\noindent
\textbf{Effectiveness of $\mathcal{L}_{\text{cont}}$.}
To show the effectiveness of our content consistency loss coupled with a content encoder, we separately feed the same source image with and without makeup to our method and show the features. 

\begin{figure}[!h]
	\centering
		\begin{minipage}[b]{0.45\linewidth}
		\centering
		\centerline{\includegraphics[width=3.8cm,height=1.9cm]{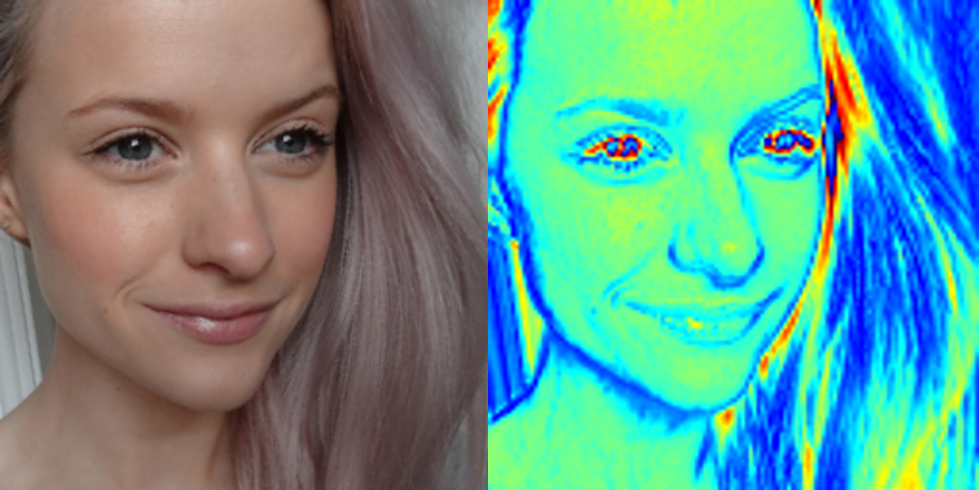}}
		\centerline{ (a) non-makeup }\medskip
	\end{minipage}
	\begin{minipage}[b]{0.45\linewidth}
		\centering
		\centerline{\includegraphics[width=3.8cm,height=1.9cm]{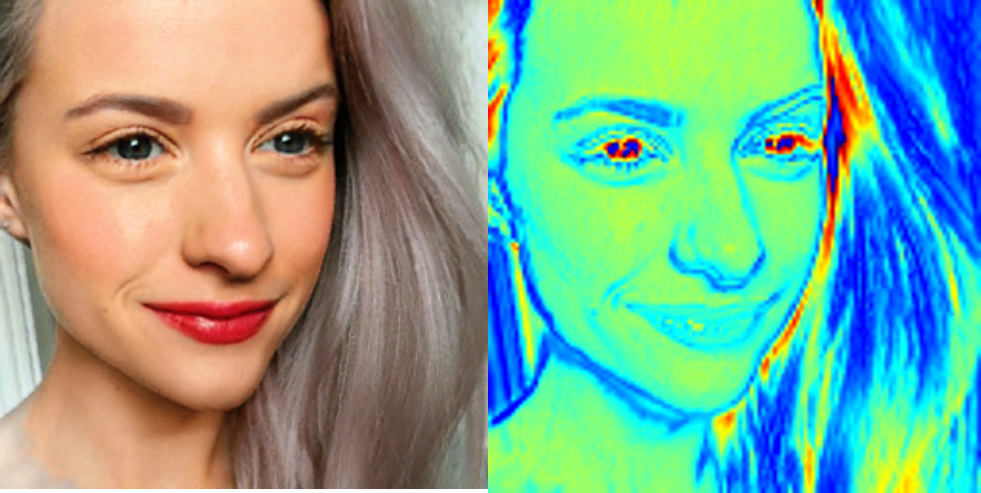}}
		\centerline{ (b) makeup}\medskip
	\end{minipage}
    \vspace{-1em}
	\caption{\textbf{Effect of the content consistency loss coupled with a content encoder.} The features are normalized and shown in heatmaps.}
	\label{fig:contconsis2}
\end{figure}

In Fig. \ref{fig:contconsis2}, we present the visualized features of the content encoder with the non-makeup image and makeup image as the input. 
With the content encoder and the proposed content consistency loss, the features of the content encoder are makeup-independent, thus benefiting the makeup transfer and avoiding the effect of makeup on the source image.

\section{Conclusion}
In this paper, we propose a makeup transfer method to overcome the limitations of previous methods such as robustness, efficiency, and the capability of content preservation. The success of our method mainly lies in the component-specific transfer together with the global transfer and the content consistency loss.  The lightweight structure and robust performance of our method outperform the state-of-the-art methods and make it suitable for practical applications. 
We also contribute a new makeup dataset, which facilitates the research of this research area.

{\small
\bibliographystyle{ieee_fullname}
\bibliography{egbib}
}

\end{document}


\title{BeautyREC: Robust, Efficient, and Content-preserving Makeup Transfer\\--Supplementary Material--}

\author{Qixin Yan$^1$\quad  Chunle Guo$^2$\quad  Jixin Zhao$^3$\quad  Yuekun Dai$^3$\quad Chen Change Loy$^3$\quad  Chongyi Li$^3$\footnotemark[1] \\
	\small{$^1$} \small WeChat, Tencent \quad
	\small{$^2$} \small TMCC, CS, Nankai University\quad	
	\small{$^3$} \small S-Lab, Nanyang Technological University \quad\\
	{\tt\small qixinyan@tencent.com \quad guochunle@nankai.edu.cn \quad zhao0388@e.ntu.edu.sg}\\
    {\tt\small  ydai005@e.ntu.edu.sg \quad ccloy@ntu.edu.sg \quad chongyi.li@ntu.edu.sg}\\
	{\tt\small \url{https://li-chongyi.github.io/BeautyREC_files}}
}
\maketitle
\thispagestyle{empty}
\renewcommand{\thefootnote}{\fnsymbol{footnote}}
\footnotetext[1]{Chongyi Li (chongyi.li@ntu.edu.sg) is the corresponding author.}

\maketitle

In this supplementary material, we present 
detailed network structure and parameters of our method and additional experimental results as follows:

\begin{itemize}
\item Details of our network structure and parameters. 
\item Discussion about the order of feature transfer.
\item Additional experimental results of our method.
\item More examples of our BeautyFace dataset.
\end{itemize}
We also discuss the broader impacts of our proposed method and dataset.
Besides, we present a video demo in a separate file. 
We  provide the code of our network in a separate file.

\section{Details of Network Structure}
In what follows, we detail our network structure and parameters, including content encoder, style encoder, component-specific correspondence, long-range dependencies, and image reconstruction.

The content encoder is composed of three convolutional layers with kernels $\{7,4,4\}$  and strides  $\{1,2,2\}$, and three resblocks. Each convolutional layer is followed by an instance norm layer and a ReLU activation function. The number of feature channels is set to 48.

The global style encoder has the same architectures and parameters as the component style encoder. The style encoder has four convolutional layers with kernels $\{7,4,4,1\}$  and strides $\{1,2,2,1\}$. The number of feature channels is set to 48. We did not adopt the normalization operation after the convolutional layers for style encoders. All layers use the ReLU as the activation function.

In the component-specific correspondence module, we use a three-layer fully-connected network with an architecture $C\rightarrow\frac{C}{r}\rightarrow C (C=48, r=16)$  to learn channel-wise weights from statistic
information of part-specific styles. For spatial attention, we separately apply mean and maximum pooling operations to the feature map that is re-weighted by the channel attention module along the channel direction. The resulting two maps by mean and maximum pooling operations are concatenated to learn a spatial attention mask using a $1\times1$ convolutional layer with the Sigmoid activation function.

In the long-range dependencies module, a transformer-based structure is employed to exploit the long-range dependencies between the source image and the reference image. The query of our model represents the global style features extracted by the global style encoder. The key and value of our model are the content features.  We use sinusoidal positional encodings  for the query and key features. The number of heads in the multi-head attention module is set to 8. A two-layer MLP with a residual connection between its input and output is placed at the end of the transformer unit.

In the image reconstruction stage, three residual blocks are  used to progressively refine features. We exploit two upsample operations and convolutional layers to recover the image resolution. The component-specific transferred features and the global transferred features are concatenated with the corresponding decoder features.
For the last convolutional layer, we use the Tanh activation function without normalization operation to produce the final result.

\section{The Order of Feature Transfer}
\begin{figure*}[!h]
	\centering
	\centerline{\includegraphics[width=0.9\linewidth]{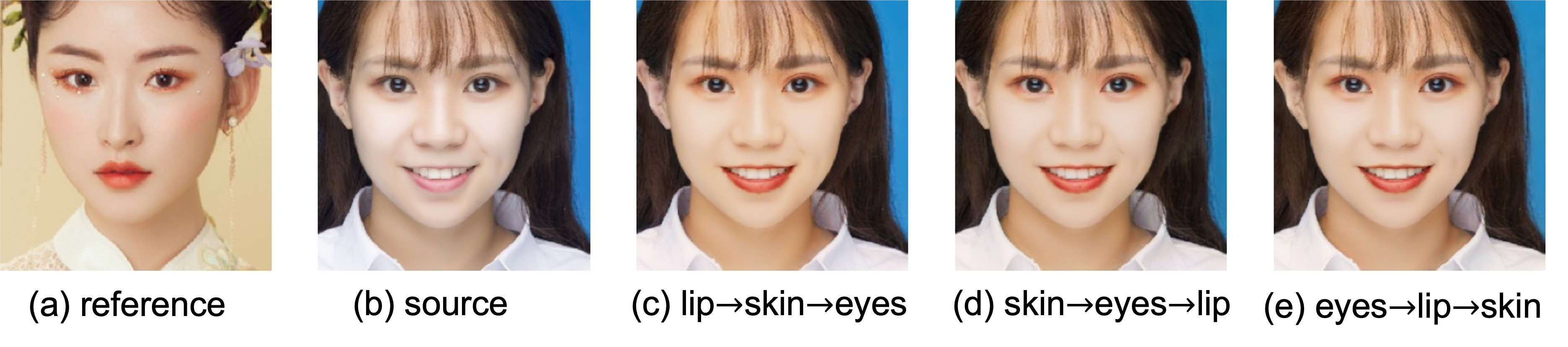}}
 \caption{The results of the different orders of feature transfer.}
 	\label{ordrer}
\end{figure*}

Our method is insensitive to the order of feature transfer. We show a set of results of the model with different component-specific transfer orders below. As shown in Fig. \ref{ordrer}, we achieve almost the same transferred results.

\section{Additional Experimental Results }
In this part, we show more experimental results of our BeautyREC as follows: 

\begin{itemize}
\item \textbf{Diversity of the results}:
Our method can generalize well across age, race, expression, etc. 
To show the generalization capability, 
we download two images of baby girls from the internet and show the makeup transfer results below. As shown, our method successfully transfers the makeup from the face of an adult female to the two baby faces. 
Besides, the visual results shown in Fig. \ref{show}. also suggest the generalization ability of our method for diverse expressions, different makeup styles, and large misalignment. 

\begin{figure*}[!h]
	\centering
	\centerline{\includegraphics[width=0.9\linewidth]{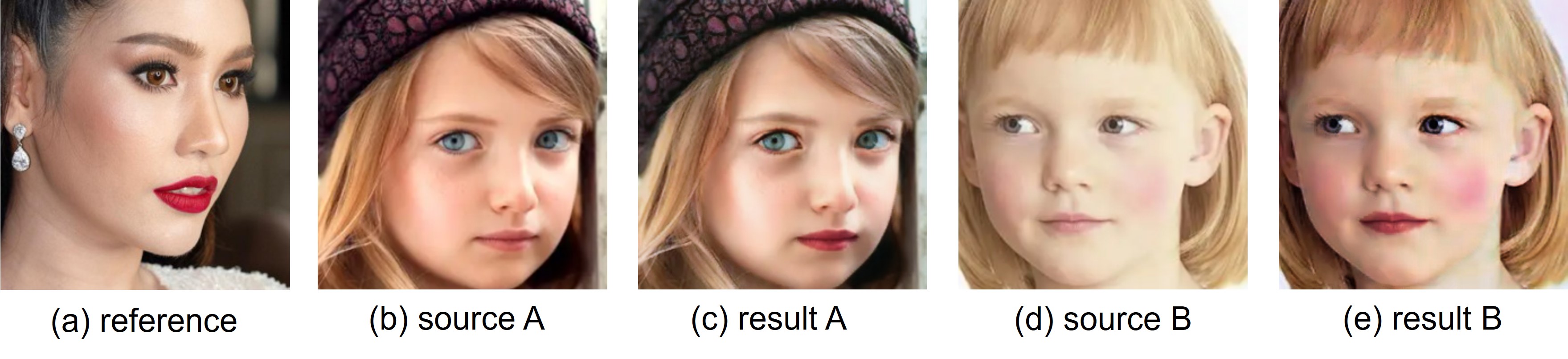}}
 \caption{The results of the different ages and expressions.}
	\label{fig2}
\end{figure*}

\item \textbf{Types of makeup:}
Our method does not simply bond the color but learns the component-specific styles guided by the reference image.
We show two more complex cases in Fig. \ref{fig3}. Our method can successfully transfer more complex cases, e.g., different lips' color and eye shadows' style.

\begin{figure*}[!h]
	\centering
	\centerline{\includegraphics[width=0.9\linewidth]{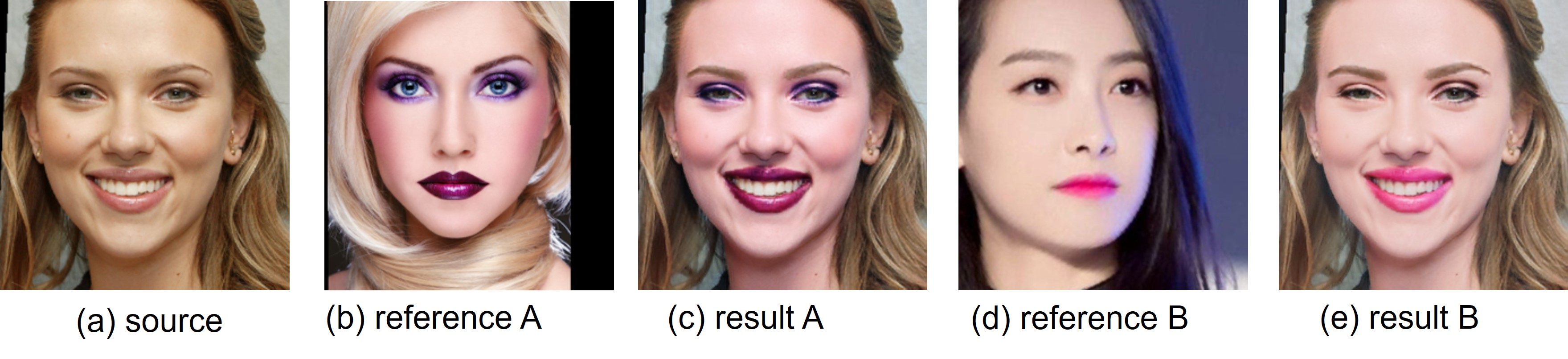}}
 \caption{The transferred results of the different types of makeup.}
 	\label{fig3}
\end{figure*}

\item \textbf{More makeup transfer results:} We present more makeup transfer results for diverse makeup styles, different backgrounds, and large misalignment in Fig. \ref{show}.

\begin{figure*}[h]
    \begin{center}
        \centering
        \includegraphics[width=0.9\textwidth]{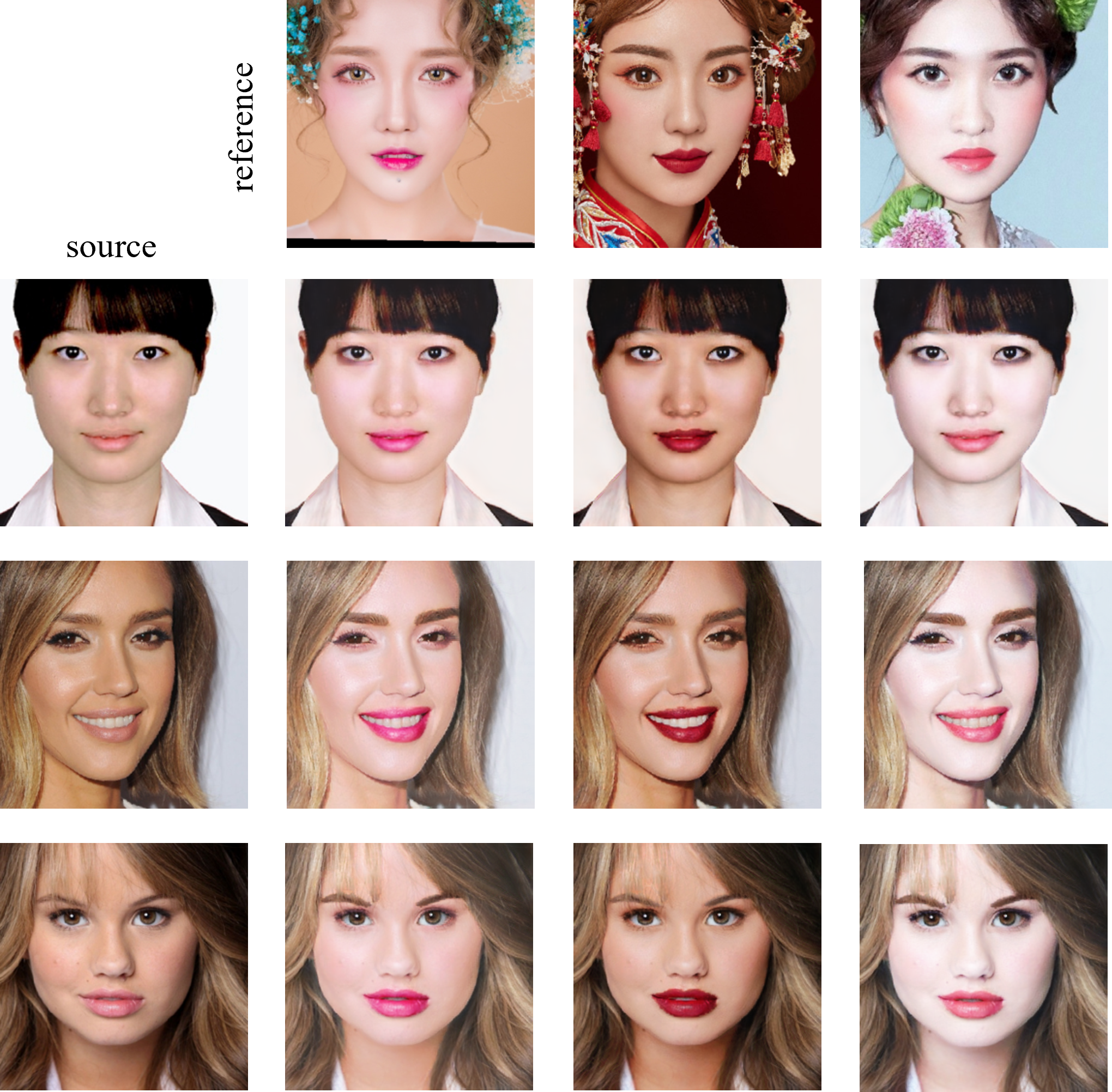}
        \captionof{figure}{\textbf{Our BeautyREC for transferring diverse makeup styles.} The source and reference images contain diverse makeup styles, different facial expressions and poses, pure backgrounds, and large misalignment.}
        \label{show}
        \vspace{-1em}
    \end{center}%
\end{figure*}

\end{itemize}

\section{More Examples of BeautyFace Dataset} 
\begin{figure*}[h]
    \begin{center}
        \centering
        \includegraphics[width=0.9\textwidth]{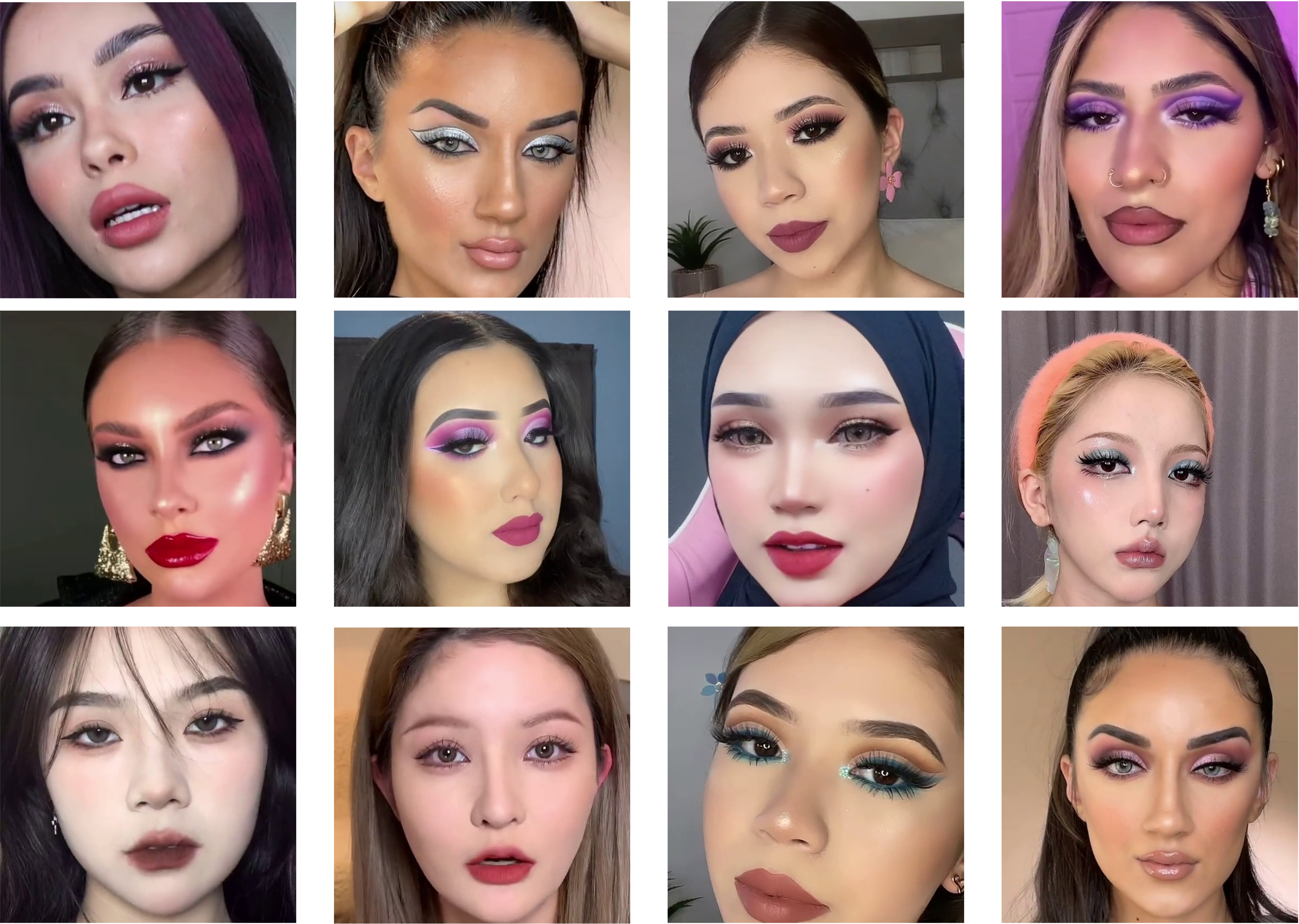}
        \captionof{figure}{\textbf{Examples of BeautyFace dataset.} BeautyFace dataset contains high-quality makeup images with diverse makeup styles, various facial expressions and poses, and large misalignment.}
        \label{face}
    \end{center}%
\end{figure*}

We provide more examples of the proposed BeautyFace dataset in Fig. \ref{face}. As Fig. \ref{face} shows, our BeautyFace dataset contains high-quality makeup images with diverse makeup styles, various facial expressions and poses, and large misalignment.

\section{Broader Impacts}
The proposed method enables efficient makeup transfer via a single-path structure, which is beneficial to the practical applications. The component-specific correspondence modeling  and long-range dependencies capturing lead to accurate and robust makeup transfer.
Besides, our high-quality BeautyFace dataset covers more recent makeup styles and more
diverse face poses, backgrounds, expressions, races, illumination, etc.
Thus, we believe our method and dataset can bring new sights and positive impacts on both academia and industry.
As a typical low-level vision work, this paper will not bring negative impacts to the society.

{\small
\bibliographystyle{ieee_fullname}
}